\documentclass[twoside]{article}

\usepackage[accepted]{aistats2021}

\usepackage[utf8]{inputenc} 
\usepackage[T1]{fontenc}    
\usepackage{url}            
\usepackage{booktabs}       
\usepackage{amsfonts}       
\usepackage{nicefrac}       
\usepackage{microtype}      
\usepackage{bm}
\usepackage{enumitem}
\usepackage{graphicx}
\usepackage{amsmath}
\usepackage{subfigure}
\usepackage[round]{natbib}
\usepackage{caption}
\usepackage{algorithm}
\usepackage{algorithmic}
\usepackage[dvipsnames]{xcolor}

\usepackage[colorlinks=true,bookmarksopen,pdfsubject={algorithms},linkcolor={MidnightBlue}, anchorcolor={black}, citecolor={MidnightBlue}, filecolor={magenta}, menucolor={black}, pagecolor={red},backref=none,urlcolor={MidnightBlue}]{hyperref}

\begin{document}

%
\runningtitle{A Scalable Gradient-Free Method for Bayesian Experimental Design  with Implicit Models}


%
\runningauthor{Jiaxin Zhang, Sirui Bi, Guannan Zhang}

\twocolumn[


\aistatstitle{A Scalable Gradient-Free Method for Bayesian \\ Experimental Design with Implicit Models}

\aistatsauthor{ Jiaxin Zhang \And Sirui Bi \And Guannan Zhang}
\aistatsaddress{ Computer Science \\ and Mathematics Division \\ Oak Ridge National Laboratory \\ \texttt{zhangj@ornl.gov} \And Computational Sciences \\ and Engineering Division  \\ Oak Ridge National Laboratory \\ \texttt{bis1@ornl.gov} \And Computer Science \\ and Mathematics Division \\ Oak Ridge National Laboratory \\ \texttt{zhangg@ornl.gov}} ]


\begin{abstract}
Bayesian experimental design (BED) is to answer the question that how to choose designs that maximize the information gathering. For implicit models, where the likelihood is intractable but sampling is possible, conventional BED methods have difficulties in efficiently estimating the posterior distribution and maximizing the mutual information (MI) between data and parameters. Recent work proposed the use of gradient ascent to maximize a lower bound on MI to deal with these issues. 
However, the approach requires a sampling path to compute the pathwise gradient of the MI lower bound with respect to the design variables, and such a pathwise gradient is usually inaccessible for implicit models. In this paper, we propose a novel approach that leverages recent advances in stochastic approximate gradient ascent incorporated with a smoothed variational MI estimator for efficient and robust BED. Without the necessity of pathwise gradients, our approach allows the design process to be achieved through a unified procedure with an approximate gradient for implicit models. Several experiments show that our approach outperforms baseline methods, and significantly improves the scalability of BED in high-dimensional problems. 
\end{abstract}

\section{Introduction}

Experimental design plays an essential role in all scientific disciplines. Our ultimate goal is to determine designs that maximize the information gathered through the experiments so that improve our understanding on model comparison or parameter estimations. A broadly used approach is Bayesian experimental design (BED) \citep{chaloner1995bayesian} that aims to find optimal design $\bm{\xi}^*$ to maximize a utility function $I(\bm \xi)$, which is typically defined by the mutual information (MI) between data and model parameters
\begin{equation}
    I(\bm \xi) = \mathbb{E}_{p(\bm{\theta},\bm{y}|\bm \xi)} \left[ \log p(\bm{\theta}|\bm{y},\bm \xi) - \log p(\bm{\theta}) \right]
\end{equation}
where $\bm \xi$ represent the experimental design, $p(\bm \theta)$ is the prior distribution of model parameters and $p(\bm \theta| \bm y, \bm \xi )$ is the posterior distribution of $\bm \theta$ given data $\bm y$ with design $\bm \xi$. However, finding $\bm{\xi}^*$ by maximizing $I(\bm{\xi})$ is a challenging task in practice because evaluating $I(\bm \xi)$ needs a nested Monte Carlo simulator, which is computationally intensive, particularly in high-dimensional settings \citep{drovandi2018improving}. 

Most of the existing BED studies focus on the {\em explicit models} \citep{chaloner1995bayesian, sebastiani2000maximum, foster2019variational,foster2020unified} where the likelihood is analytically known but in nature and science, the more common scenario is the {\em implicit models} \citep{kleinegesse2019efficient,overstall2018bayesian}, where the likelihood is intractable and not evaluated directly but sampling is possible. In other words, the model is specified based on a stochastic data generating simulator and typically has no access to the analytical form and the gradients of the joint density $p(\bm{\theta},\bm{y}|\bm \xi)$ and marginal density $p(\bm y|\bm \xi)$. The resulting BED scheme typically shares a two-stage feature: build a pointwise estimator of $I(\bm \xi)$ and then feed this ``black-box'' estimator to a separate outer-level optimizer to find the optimal design $\bm{\xi}^*$. This framework substantially increases the overall computational cost and is difficult to scale the BED process to a high dimensional design space. Recent studies \citep{kleinegesse2020bayesian, foster2020unified,harbisher2019bayesian} alleviate the challenges by introducing stochastic gradient-based approaches but they rely on the models with tractable likelihood functions or assume the gradient can be reasonably approximated by pathwise gradient estimators with sampling path \citep{kleinegesse2020bayesian}, which requires that we can sample from the data distribution $p(\bm y | \bm \theta, \bm \xi)$ by sampling from a base distribution $p(\bm \epsilon)$ and then transforming the samples through a specific function $g(\bm \epsilon; \bm \theta, \bm \xi)$, which is called the sampling path, i.e., $\bm y \sim p(\bm y | \bm \theta, \bm \xi) \Longleftrightarrow \bm y = g(\bm \epsilon; \bm \theta, \bm \xi), \bm \epsilon \sim p(\bm \epsilon)$. This is unlike the scope of our paper that focuses on the BED for implicit models without gradients. 

\subsection{Related Work}
\citet{foster2019variational} recently proposed to use a lower bound of MI for BED. This study relies on variational approximations to the likelihood and posterior but it is a two-stage approach where the optimal designs were determined by a separate Bayesian optimization (BO). Unfortunately, this approach has a limitation in scaling to high-dimensional design problems. A follow-up study developed by \cite{foster2020unified} aims to address the scalability issue by introducing a unified stochastic gradient-based approach. However, they assumed the models with the tractable explicit likelihood or gradient approximations are available. In the scope of BED for implicit models, \citet{ao2020approximate} proposed an approximate KLD based BED method for models with intractable likelihoods; \citet{kleinegesse2019efficient, kleinegesse2020sequential} have recently considered the use of MI combined with likelihood-free inference by ratio estimation to approximate the posterior distribution but this method is often computationally intensive. The authors rectify this in a follow-up study \citep{kleinegesse2020bayesian} that leverages mutual information neural estimation (MINE) \citep{belghazi2018mine} to jointly determine the optimal design and the posterior distribution. However, for the experimental design without gradients, this method falls back to the gradient-free methods, i.e., BO at the expense of reduced scalability, to solve the optimization problems within the two-stage framework. Moreover, the variational BED methods \citep{kleinegesse2020bayesian, foster2019variational,foster2020unified} may yield an unstable design with a high variance on the posterior distribution due to the use of variational MI estimators that exhibit a high variance argued by \citet{song2019understanding}. 

{\bf Contributions} \quad In this paper, we propose a novel Bayesian experimental design approach for implicit models without available gradients. This scalable method can address the aforementioned technical challenges, particularly in high dimensional problems. 

\begin{itemize}[leftmargin=10pt]
\item We propose a general unified framework that leverages stochastic approximate gradient without the requirement or assumption of pathwise gradients and sampling paths for implicit models.
\item We propose to use a smoothed MI lower bound to conduct robust MI estimation and optimization, which allows the variance of the design and posterior distribution to be much lower than existing approaches. 
\item We show the superior performance of the proposed approach through several experiments and demonstrate that the approach enables the optimization to be performed by a stochastic gradient ascent algorithm and thus well scaled to considerable high dimensional design problems. 
\end{itemize}

\section{Bayesian Experimental Design}
The Bayesian experimental design (BED) framework aims at choosing an experimental design $\bm \xi$ to maximize the information gained about some parameters of interest $\bm \theta$ from the outcome $\bm y$ of the experiment. Typically, the BED framework begins with a Bayesian model of the experimental process, including a prior distribution $p(\bm \theta)$ and a likelihood $p(\bm y|\bm \theta,\bm \xi)$. The information gained about $\bm \theta$ from running the experiment with design $\bm \xi$ and observed outcome $\bm y$ can be interpreted by the reduction in entropy from the  prior to posterior 
\begin{equation}
    \textup{IG}(\bm y, \bm \xi) = \mathcal{Q}[p(\bm \theta)] - \mathcal{Q}[p(\bm \theta | \bm y, \bm \xi)].
\end{equation}
To define a metric to quantify the utility of the design $\bm \xi$ before running experiments, an expected information gain (EIG), $I(\bm \xi)$ is often used by
\begin{equation}
    I(\bm \xi) = \mathbb{E}_{p(\bm y|\bm \xi)}[\mathcal{Q}[p(\bm \theta)] - \mathcal{Q}[p(\bm \theta | \bm y, \bm \xi)]]. \label{eq:eig}
\end{equation}
Eq.~\eqref{eq:eig} can also be interpreted as a mutual information (MI) between $\bm \theta$ and $\bm y$ with a specified $\bm \xi$, 
\begin{equation}
    I_{\rm MI} (\bm \xi) = \mathbb{E}_{p(\bm \theta)p(\bm y|\bm \theta, \bm \xi)} \left[ \log\frac{p(\bm y|\bm \theta, \bm \xi)}{p(\bm y | \bm \xi)} \right]. \label{eq:BED_MI}
\end{equation}
The BED problem is therefore defined as 
\begin{equation}
    \bm \xi^* = \operatorname*{arg\,max}_{\bm \xi \in \bm \Xi} I_{\rm MI} (\bm \xi), \label{eq:max_mi}
\end{equation}
where $\bm \Xi$ is the feasible design domain. The most challenging task in the BED framework is how to efficiently and accurately estimate $I_{\rm MI} (\bm \xi)$ in Eq.~\eqref{eq:BED_MI} and optimize $I_{\rm MI} (\bm \xi)$ via Eq.~\eqref{eq:max_mi} to obtain the optimal design $\bm \xi^*$. The following section will discuss the two core tasks in the BED framework, that is MI estimation and optimization. 

\subsection{Mutual Information Estimation}
Mutual information (MI) estimation plays a critical role in many important problems, not only the BED framework but also other machine learning tasks such as reinforcement learning \citep{pathak2017curiosity} and representation learning \citep{chen2016infogan,oord2018representation}. However, estimating mutual information from samples is always challenging \citep{mcallester2020formal}. This is because that, in general, neither $p(\bm \theta | \bm y, \bm \xi)$ and $p(\bm y | \bm \xi)$ have analytical closed-form so that classical Monte Carlo methods are intractable to compute the integral, specifically in high dimensions. One potential approach is to use a nested Monte Carlo (NMC) estimator \citep{myung2013tutorial,rainforth2018nesting}, which is given by 
\begin{equation}
    I_{\rm NMC}(\bm \xi) = \frac{1}{N}\sum_{i=1}^{N} \log\left[\frac{p(\bm y_i | \bm \theta_{i,0},\bm \xi)}{\frac{1}{m}\sum_{j=1}^M p(\bm y_i|\bm \theta_{i,j}, \bm \xi)}\right],
\end{equation}
where $\bm \theta_{i,j} \sim p(\bm \theta)$ are 
independent and identically distributed (i.i.d) samples, and $\bm y_{i} \sim p(\bm y | \bm \theta = \bm \theta_{i, 0}, \bm \xi)$. However, NMC estimator requires both the inner and outer integrals and therefore converges slowly at an overall rate of $\mathcal{O}(T^{-1/3})$ \citep{rainforth2018nesting} in the total computational cost $T=\mathcal{O}(NM)$. 

Recently, there has been an increasingly interest in MI estimation with variational methods \citep{barber2003algorithm,nguyen2010estimating,foster2019variational}, which can benefit from a natural incorporation with deep learning algorithms \citep{alemi2016deep,oord2018representation,belghazi2018mine,poole2019variational}. 

Although the variational approaches to MI estimation have been widely used, there are still several limitations pointed out by \citet{song2019understanding}, for example, a high variance of the estimators in MINE \citep{belghazi2018mine}. This will bring additional challenges for the subsequent MI optimization. To mitigate the issues, \citet{song2019understanding} proposed a unified framework over variational estimators that treat variational MI estimation as an optimization problem over density ratios. This is achieved by utilizing the role of partition function estimation with variance reduction techniques. This improved MI estimation, named by a {\em smoothed MI lower bound estimator}, $I_{\rm SMILE}$, has been demonstrated by compared to existing variational methods in \citet{barber2003algorithm, poole2019variational}.

\subsection{Mutual Information Optimization}
The BED problem is to find the optimal design that maximizes the MI estimation. Given a point-by-point base MI estimator, a variety of different approaches can be used for the subsequent optimization over designs \citep{amzal2006bayesian,muller2005simulation, rainforth2017automating}. At a high-level, most existing methods belong to a two-stage procedure where an MI estimator of $I_{\rm MI} (\bm \xi)$ is first made and then followed by a separate optimization algorithm which is used to select the candidate design $\bm \xi$ to evaluate next. However, this two-stage framework can be computationally intensive because it needs an extra level of nesting to the optimization process in Eq.~\eqref{eq:max_mi}; in other words, $I_{\rm MI}(\bm \xi)$ must be separately estimated for each $\bm \xi$, which significantly increase the overall computational cost. 

Recent studies \citep{foster2019variational,kleinegesse2019efficient} tend to use Bayesian optimization (BO) \citep{snoek2012practical} for this purpose due to its sample efficiency, robustness and capability to naturally deal with noisy MI estimator \citep{foster2019variational}. However, it is difficult for BO to scale the overall BED process to high-dimensional design settings \citep{foster2020unified, kleinegesse2020bayesian}. Broadly speaking, BO is prohibitively slow and approaches the performance of ``random search'' if the dimensionality of design space is $\Omega$>100 \citep{snoek2015scalable}. The scalability of BO is still the major challenge, even though some recent studies \citep{wang2017batched,li2018high,mutny2018efficient,rana2017high,li2016high,eriksson2019scalable} are involved to make improvements.   

To open the door of high-dimensional BED, \citet{foster2019variational} proposed a unified gradient-based BED method but they focus on models with tractable likelihood functions or assume that the gradients are available. \citet{kleinegesse2020bayesian} proposed a MINEBED method that leverages the neural MI estimation to jointly determine the optimal design and the posterior using a gradient-based optimization algorithm. However, the MINEBED only works for the experimental design with available pathwise gradient estimators. As for the implicit models without gradient, \citet{kleinegesse2020bayesian} still consider a two-stage framework that uses BO as the outer-level optimizer at the expense of reduced scalability.

\section{The SAGABED Approach}
We here show how to perform a stochastic approximate gradient ascent (SAGA) method to design optimal experiments for implicit models without gradients and address the computational challenges in scaling to high-dimensional problems. 

\subsection {Smoothed MI Estimator} 
\citet{belghazi2018mine} have recently proposed to estimate MI by gradient ascent over neural networks and argued that the lower bound can be tightened by optimizing the neural network parameters. This MI estimator is typically named by MINE-$f$ or $f$-GAN KL \cite{nowozin2016f}
\begin{equation}
\begin{aligned}
    I_{\rm MINE}(\bm \xi, \bm \psi) = &\; \mathbb{E}_{p(\bm{\theta},\bm{y}|\bm \xi)}[\mathcal{T}_{\bm \psi}(\bm \theta, \bm y)] \\  & -\log\left( \mathbb{E}_{p(\bm{\theta})p(\bm y|\bm \xi)}\left[e^{\mathcal{T}_{\bm \psi}(\bm \theta, \bm y)}\right]\right), \label{eq:mine}
\end{aligned}
\end{equation}
where $\mathcal{T}_{\bm \psi}(\bm \theta, \bm y)$ is a neural network that is parametrized by $\bm \psi$ with model parameter $\bm \theta$ and data $\bm y$ as input. Incorporating neural network parameters $\bm \psi$ with design parameters $\bm \xi$, the BED problem can be formulated by maximizing the overall objective
\begin{equation}
    \bm{\xi}^* = \operatorname*{arg\,max}_{\bm \xi}\max_{\bm \psi} \left\{I_{\rm MINE}(\bm \xi, \bm \psi) \right\}. \label{eq:obj}
\end{equation}
The optimal design $\bm{\xi}^*$ can be obtained by maximizing the MI estimator in Eq.~\eqref{eq:mine} through a joint gradient-based algorithm \citep{kleinegesse2020bayesian,foster2020unified} or a separate gradient-free updating scheme \citep{foster2019variational,kleinegesse2020bayesian} of $\bm \xi$ and $\bm \psi$. The accuracy, efficiency and robustness for estimating and optimizing MI therefore become very important to the BED tasks. Unfortunately, $I_{\rm MINE}$ exhibits variance that could grow exponentially with the ground truth MI and leads to poor bias-variance trade-offs in practice \citep{song2019understanding}. The high variance of $I_{\rm MINE}$ may weaken the robustness the final optimal design $\bm \xi^*$ and thus cause a higher variance of the posterior distributions. We propose to use a smoothed mutual information lower-bound estimator $I_{\rm SMILE}$ with hyperparameter $\tau$ \citep{song2019understanding} 
\begin{equation}
\begin{aligned}
    & I_{\rm SMILE}(\bm \xi, \bm \psi) \\ = & \;\mathbb{E}_{p(\bm{\theta},\bm{y}|\bm \xi)}[\mathcal{T}_{\bm \psi}(\bm \theta, \bm y)] \\ & -\log\left(\mathbb{E}_{p(\bm{\theta})p(\bm y|\bm \xi)}[{\rm clip}(e^{\mathcal{T}_{\bm \psi}(\bm \theta, \bm y)}, e^{-\tau}, e^{\tau})]\right), \label{eq:smile}
\end{aligned}
\end{equation}
where the clip function is defined as 
\begin{equation}
    \textup{clip} (u,v,w) = \max(\min(u,w), v).
\end{equation}
The choice of $\tau$ affects the bias-variance trade-off: when $\tau \rightarrow \infty$, $I_{\rm SMILE}$ converges to $I_{\rm MINE}$; with a smaller $\tau$, the variance is reduced at the cost of increasing bias \citep{song2019understanding}. The improved MI estimation via variance reduction techniques is benefit to the optimizing process in Eq.~\eqref{eq:obj} and thus yields a robust optimal design $\bm{\xi}^*$. 

\subsection{Stochastic Gradient Approximate} 
In the context of BED for implicit models, the pathwise gradients of the MI lower bound are unavailable. A potential choice for approximating the gradient is to use Gaussian Smoothing (GS) \citep{nesterov2017random} method to make the function smooth. The smoothed loss is defined by
\begin{equation}
    f_{\sigma}(\bm \xi) = \mathbb{E}_{\bm \epsilon \sim \mathcal{N}(0, \mathbf{I}_d)} \left[f(\bm \xi + \sigma \bm \epsilon) \right], \label{eq:GS}
\end{equation}

where $\mathcal{N}(0, \mathbf{I}_d)$ is the $d$-dimensional standard Gaussian distribution, and $\sigma > 0$ is the smoothing radius. The standard GS represents the $\nabla f_\sigma(\bm \xi)$ as an $d$-dimensional integral and estimate it by drawing $M$ random samples $\{\bm \epsilon_i\}_{i=1}^M$
from $\mathcal{N}(0,\mathbf{I}_d)$, i.e., 
\begin{equation} \label{eq:es_grad} 
\begin{aligned}
    \nabla f_{\sigma}(\bm \xi) & = 
    \frac{1}{\sigma}\mathbb{E}_{\bm \epsilon \sim \mathcal{N}(0, \mathbf{I}_d)} \left[f(\bm \xi + \sigma \bm \epsilon)\, \bm \epsilon \right] \\& \approx \frac{1}{M\sigma}\sum_{i=1}^M f(\bm \xi + \sigma \bm \epsilon_i)\bm \epsilon_i.
\end{aligned}
\end{equation}
This method using random MC sampling is also called Evolution Strategies (ES) which have been successfully applied in a variety of high-dimensional optimization problems, including RL tasks \citep{salimans2017evolution, choromanski2018structured} due to the advantages in scalability and parallelization capability. However, gradient estimator of ES tends to have a higher variance for high-dimensional space such that it requires a large number of samples to be robust \citep{nesterov2017random}. To reduce the variance of the gradient estimator and incorporate with the smoothed MI estimator $I_{\rm SMILE}$, we 
propose to use the Guided ES (GES) algorithm that leverages the recent advances in variance-reduced ES \citep{maheswaranathan2018guided}. Specifically,  GES generates a subspace by keeping track of the previous $k$ surrogate gradients during optimization, and leverage this prior information by changing the distribution of $\bm \epsilon_i$ in Eq.~\eqref{eq:es_grad} to $\mathcal{N}(0, \bm \Sigma)$ with
\begin{equation}
    \bm \Sigma = (\alpha/n) \cdot \mathbf{I}_n + (1-\alpha)/k \cdot UU^T, \label{eq:GES}
\end{equation}
where $k$ and $n$ are the subspace and parameter dimensions respectively, $U$ denotes an $n \times k$ orthonormal basis for the subspace, and $\alpha$ is a hyperparameter that trades off variance between the subspace and full parameter space. This improved search distribution allows a low-variance estimate of the descent direction $\nabla f_{\sigma}^G(\bm \xi)$, which can then be passed to a stochastic gradient ascent (SGA) optimizer. The GES algorithm is further improved by utilizing historical estimated gradients to build a low-dimensional subspace for sampling search directions and adaptively update the importance of this subspace through a exploitation and exploration trade-off \citep{liuself}. To this end, the variance of the GES gradient estimator $\hat{\nabla} f_{\sigma}^G(\bm \xi)$ is low and it can be naturally incorporated with $I_{\rm SMILE}$ to find a robust optimal design $\bm \xi^*$. 

\subsection{SAGABED Algorithm for Implicit Models}

To address the grand challenges in high-dimensional BED for implicit models without gradients, we propose a novel and scalable approach, named by \texttt{SAGABED}, which integrates a smoothed MI lower bound and an approximate gradient estimator. The details of \texttt{SAGABED} algorithm is provided in Algorithm \ref{algo:1}. 

\begin{algorithm}[h!]
  \caption{\hspace{-0.1cm}: The \texttt{SAGABED} algorithm}
\begin{algorithmic}[1] \label{algo:1}
\STATE{\bf Require}: neural network architectures, learning rates $\ell_{\psi}$ and $\ell_{\xi}$, $\tau$ in $I_{\rm SMILE}$, total prior samples $n$, total iterations $T$, implicit model $\mathcal{M}$
\STATE{\bf Process}:
\STATE Initialize a design $\bm \xi_0$ by random sampling
\STATE Initialize neural network parameter $\bm \psi_0$
\FOR{$t=0:T-1$}
\STATE Draw $n$ samples from the prior distribution of the model parameters $\bm \theta$: \\
$ \bm \theta^{(1)},...,\bm \theta^{(n)} \sim p(\bm \theta) $
\STATE Compute the corresponding data samples $\bm y^{(i)}$, $i=1,...,n$ using the current design $\bm \xi_t$ and an implicit model $\mathcal{M}$
\STATE Evaluate the smoothed MI lower bound $I_{\rm SMILE}$ by Eq.~\eqref{eq:smile} at the current design $\bm \xi_t$ and network parameters $\bm \psi_t$
\STATE Compute the approximate gradient estimator $\nabla_{\bm \xi}^* I_{\rm SMILE} (\bm \xi_t, \bm \psi_t)$ using the GES algorithm
\STATE Evaluate the gradient of the $I_{\rm SMILE}$ with respect to the network parameters $\nabla_{\bm \psi} I_{\rm SMILE} (\bm \xi, \bm \psi)$
\STATE Update design $\bm \xi_t$ via gradient ascent: \\
$\bm \xi_{t+1} = \bm \xi_{t} + \ell_{\xi} \nabla_{\bm \xi}^* I_{\rm SMILE} (\bm \xi_t, \bm \psi_t) $
\STATE Update neural network parameters $\bm \psi_t$ via gradient ascent: \\
$\bm \psi_{t+1} = \bm \psi_{t} + \ell_{\psi} \nabla_{\bm \psi} I_{\rm SMILE} (\bm \xi_t, \bm \psi_t) $
\ENDFOR
\end{algorithmic}
\end{algorithm}

In the following, we discuss some important features of the \texttt{SAGABED} approach, specifically for high-dimensional design problems. 

\begin{itemize}[leftmargin=12pt]
    \item { {\em Unified framework vs two-stage framework}} \\  Without the requirement of pathwise gradients for implicit models, we utilize the stochastic approximate gradients and construct a unified framework that allows the design process to be performed by a simultaneous optimization with respect to both the variational and design parameters.  The existing two-stage framework that builds a pointwise MI estimator before feeding this estimator to an outer-level optimizer is often computationally intensive. 
    \item {\em Scalability, portability, and parallelization} \\ we proposes the stochastic approximate gradient ascent procedure that naturally avoids to the scalability issue in gradient-free methods including BO. The proposed framework can be easily incorporated with other MI estimators and implicit models because we only need the forward simulation value to approximate the gradient using guided ES algorithm. Compared to gradient-based methods, our computational cost is slightly higher but we benefit from the parallelization capability of Guided ES methods and thus improve the computational efficiency. 
    \item {\em Robust estimation with a lower variance}\\ The smoothed MI lower bound used in this method allows us to perform a robust MI estimation and optimization for Bayesian experimental design. The resulting low variance of the optimal design and posterior samples enable a more preciously estimate of the model parameters.   
\end{itemize}
After determining the optimal design $\bm{\xi}^*$ by maximizing the MI lower bound, we can obtain an estimate of the posterior $p(\bm \theta | \bm y,  \bm{\xi}^*)$ given the learned neural network $\mathcal{T}_{\bm{\psi}^*}(\bm \theta, \bm y)$ and prior distribution
\begin{equation}
    p(\bm \theta | \bm y, \bm \xi) = {\rm clip}(e^{\mathcal{T}_{\bm \psi}(\bm \theta, \bm y)-1}, e^{-\tau}, e^{\tau}) p(\bm \theta). \label{eq:posterior}
\end{equation}
The relationship in Eq.~\eqref{eq:posterior} allows to easily generate posterior samples $\bm \theta_i \sim p(\bm \theta | \bm y, \bm \xi^*)$ using Markov chain Monte Carlo (MCMC) sampling or categorical sampling \citep{kleinegesse2020bayesian} since the posterior density can be quickly evaluated via Eq.~\eqref{eq:posterior}. 

\section{Experiments}
We here demonstrate our method in three examples: a linear model with several noises, a real-world pharmacokinetic model \citep{ryan2014towards}, and a scientific quantum control problem \citep{nistobed}. We compare \texttt{SAGABED} against two baselines: the two-stage framework using BO and the gradient-based framework (if gradient is available). The first two have sampling path gradients, allowing us to compare with gradient-based methods, while the third one does not. 

\subsection{ Noisy Linear Regression}

We first show our approach through a classical linear model with noisy sources that has been used by \citet{kleinegesse2020bayesian} and \cite{foster2020unified}. We assume the model to be  
\begin{equation}
    \bm y= \theta_1 \bm 1 + \theta_2 \bm \xi + \bm \epsilon + \bm \nu, \label{eq:linear}
\end{equation}
where $\bm y$ is response variables, $\bm \theta=[\theta_1, \theta_2]^T$ are model parameters, $\bm \epsilon \sim \mathcal{N}(0,1)$ and $\bm \nu \sim \Gamma(2,2)$  are noise terms. The design problem is to make $D$ measurements to better estimate the model parameters $\bm \theta$ by constructing a design vector $\bm \xi = [\xi_1,...,\xi_D]^T$ which consists of individual experimental design. Using the linear model in Eq~\eqref{eq:linear}, we can obtain the corresponding independent measurement $\bm y_i$, which gives a data vector $\bm y = [y_1, ...,y_D]^T$.

This toy example enables us to use numerical integration to approximate the posterior and MI but here we assume it is an implicit model for testing our method. Since the sampling path is given by Eq~\eqref{eq:linear}, we can easily compute the pathwise gradients with respect to the designs $\nabla_{\bm \xi} I(\bm \xi, \bm \psi) $ in \citet{kleinegesse2020bayesian} and thus find the optimal designs $\bm \xi^* = [\xi_1^*, ...,\xi_D^*]$ using gradient-based approach, which is chosen as a baseline for comparison. 

We start from a simple case with only one measurement, i.e., $D=1$. The initial design is randomly drawn from the design domain $\bm \xi \in [-10, 10]$ and 10,000 samples of model parameters $\bm \theta$ are generated from the the prior distribution $p(\bm \theta) = \mathcal{N}(0, 3^2)$. To estimate the MI lower bound, we use a neural network $\mathcal{T}_{\bm \psi} (\bm \theta, \bm y)$ with one layer of 100 neurons, with a \texttt{ReLU} activation function, as well as \texttt{Adam} optimizer with learning rates $\ell_{\psi}=10^{-4}$ and $\ell_{\xi} = 10^{-2}$. The choice of $\tau$ in $I_{\rm SMILE}$ affects the bias-variance trade-off, so we compared $\tau=1,5,10$ and selected the $\tau=5$ with a smaller variance but an acceptable bias for all examples.

\begin{figure}[h!]
    \centering
    \includegraphics[width=0.22\textwidth]{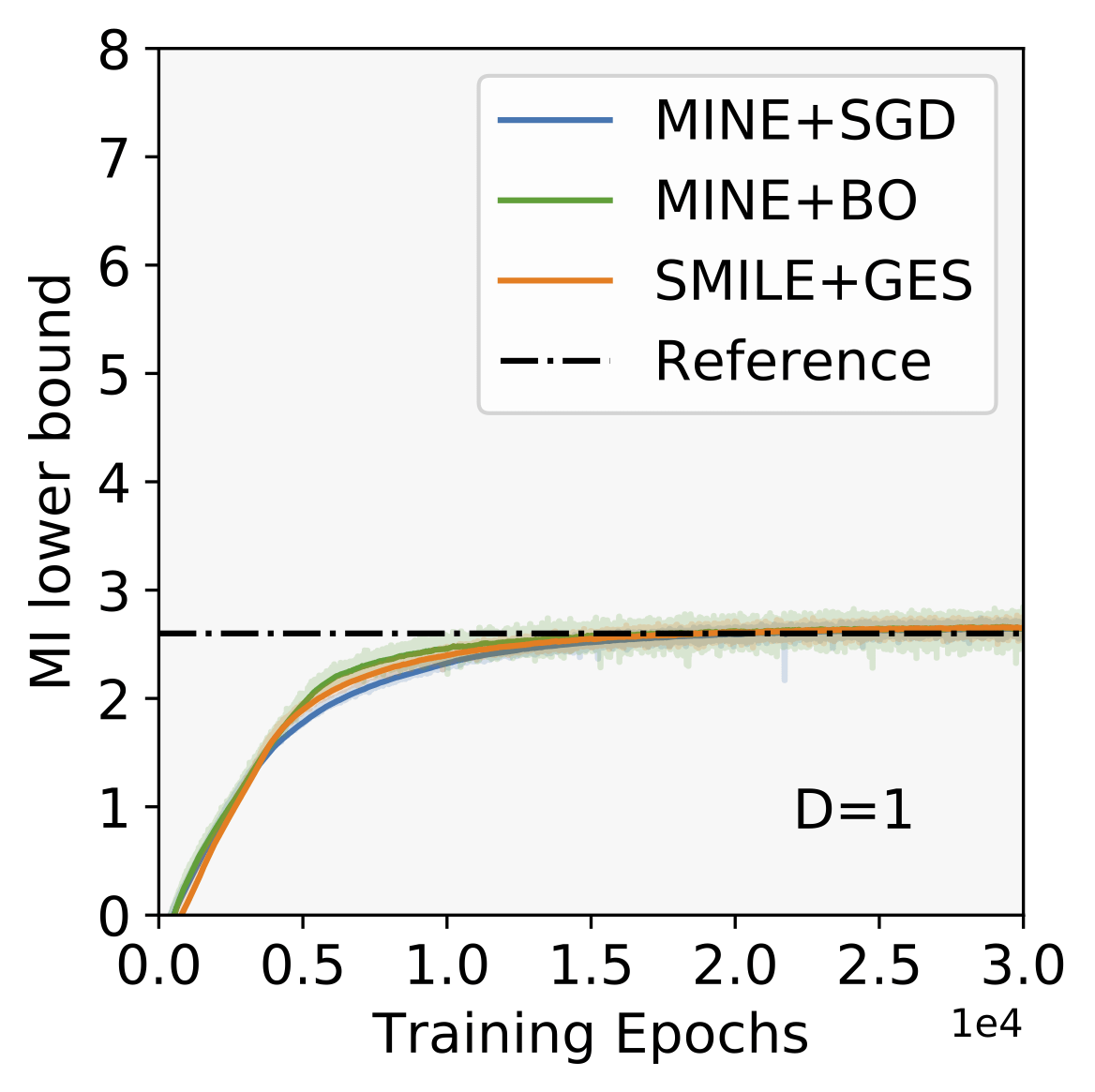}
    \includegraphics[width=0.22\textwidth]{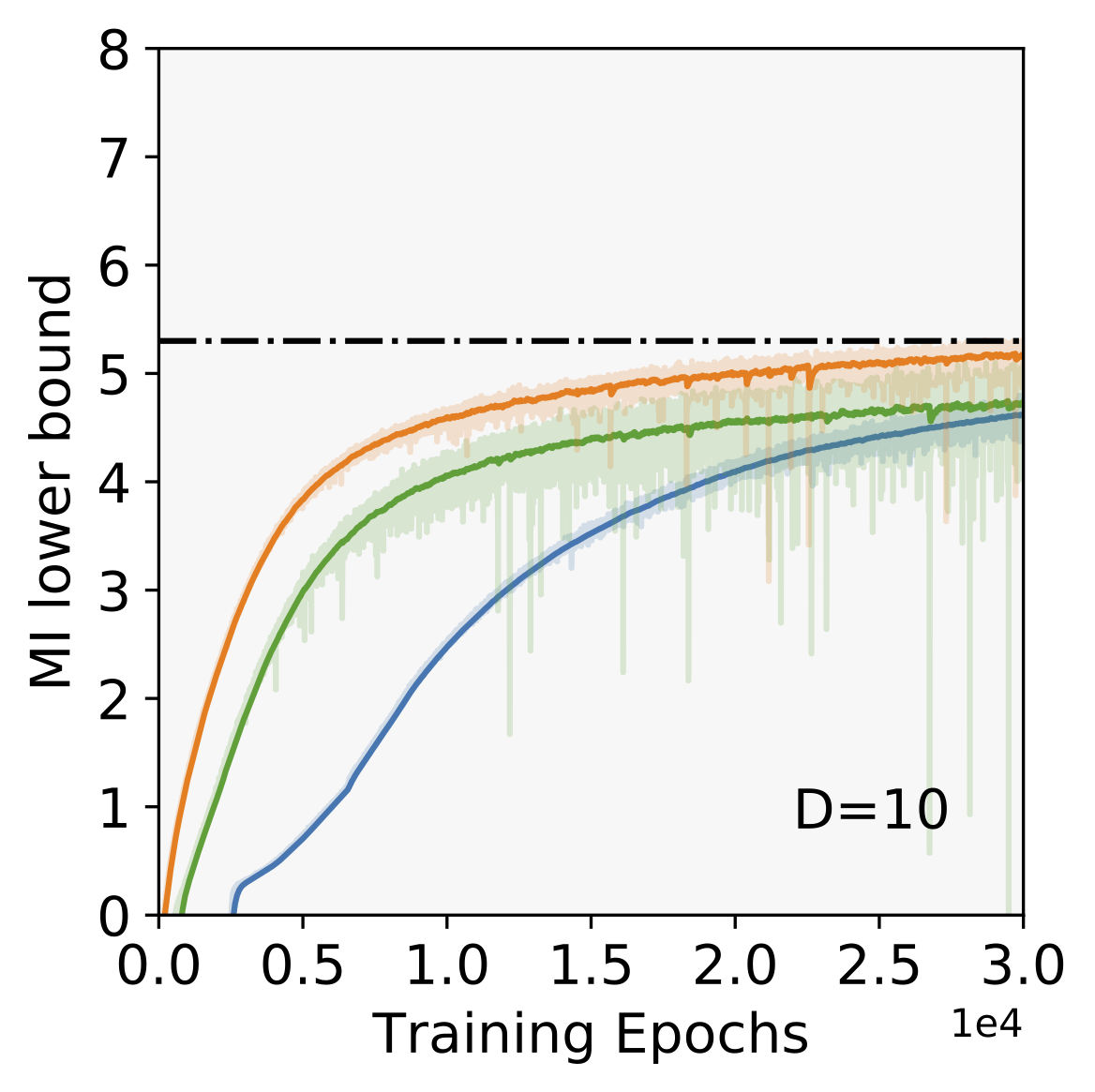}
    \includegraphics[width=0.22\textwidth]{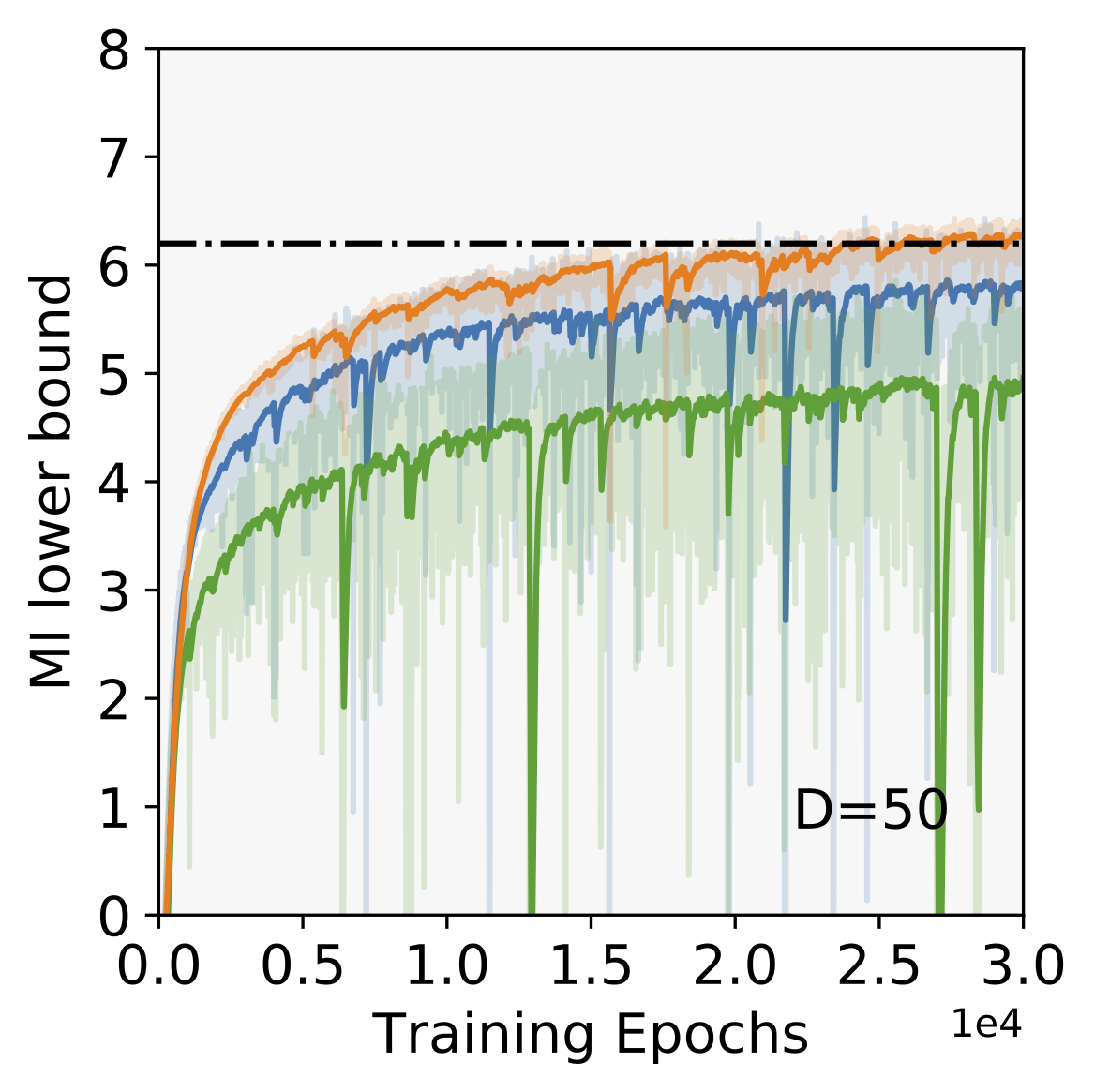}
    \includegraphics[width=0.22\textwidth]{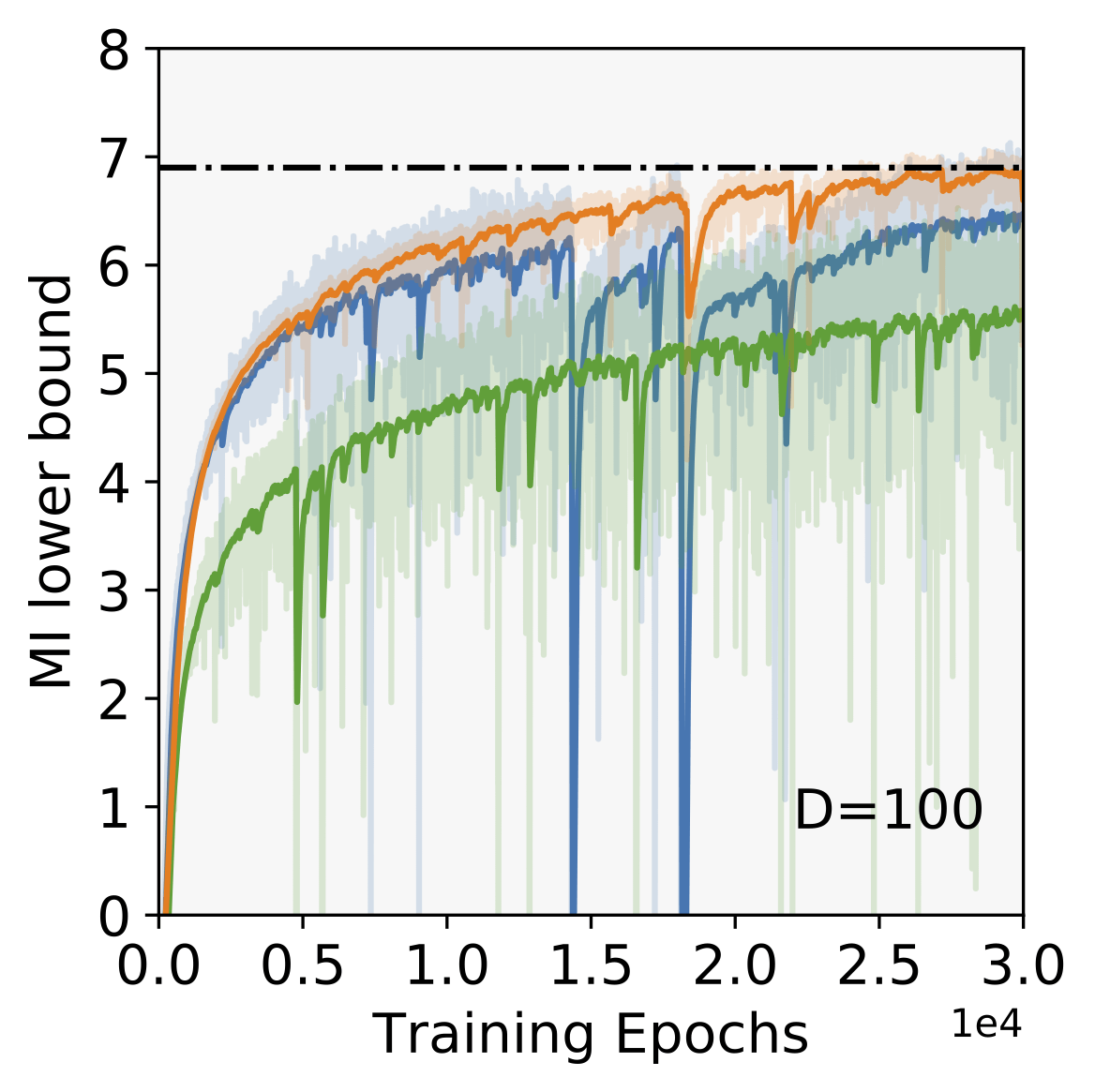}
    \caption{The MI lower bound as a function of neural network training epochs for $D$=1, 10, 50 and 100 measurements in noisy linear model. The dotted lines are reference MI values at optimized design $t^*$ computed by the nested MC approach.}
    \label{fig:noise}
\end{figure}

Figure \ref{fig:noise} shows the MI lower bound as a function of neural network training epochs. Except for the proposed \texttt{SAGABED} method that integrates $I_{\rm SMILE}$ and Guided ES algorithm, there are two baselines for comparison: $I_{\rm MINE}$ with SGD (blue curve) and $I_{\rm MINE}$ with BO (green curve). When $D=1$, three methods show a fast convergence of the MI lower bound that is around 2.6, close to a reference MI value computed by the nested MC method and the computing details can be found in the supplement material. The final optimal design found by three methods are all at the boundary, i.e., $\xi^*=$ 10 or -10, which is intuitive due to a larger signal-to-noise ratio at designs for the linear model. 

For more complex cases, e.g., high-dimensional design problems, we adjust the neural network architectures by hyperparameter optimization and determine to use one layer with 150 neurons for $D=10$ and five layers with 50 neurons in each layer for $D=$50 and 100. We use the same activate function, learning rate, and optimizer as before. As shown in Figure \ref{fig:noise}, the final MI lower bound estimated by \texttt{SAGABED} has a bias to the reference MI values but negligibly small compared to the baselines. The gradient-based method shows a comparative performance as \texttt{SAGABED} but it is not stable and tends to collapse during training, e.g., $D=100$ in Figure \ref{fig:noise}, in the meantime, it might be difficult to escape from a local minimum, which may yield suboptimal results, particularly in high-dimensional design space. The method using BO is limited by its computational and scalable issues, and shows a large bias compared to the reference MI value at $D=50$ and 100. We also demonstrate superior performance of the \texttt{SAGABED} on estimation variance that is much smaller than the baselines across all tasks shown in Figure \ref{fig:noise}. 

After finding the optimal design $\bm \xi^*$, we can compute the corresponding data $\bm y^*$ by performing a real-world experiment; then we can estimate the model parameters $\bm \theta$. Here we assume to have a true model parameter $\bm \theta_{\rm true} = [1,4]$ and then use it to generate $\bm y^*$. We can compute the posterior density using the trained neural network $\mathcal{T}_{\bm \psi^*}(\bm \theta, \bm y)$ and Eq.~\eqref{eq:posterior}. We therefore obtain the posterior samples from MCMC sampling and use them to estimate the model parameters. Table \ref{tab:noise} shows the estimating mean and standard deviation of the posterior samples of the model parameters $\bm \theta = (\hat{\theta}_1, \hat{\theta}_2)$. The estimating error in $D=1$ is very large because only one measurement is naturally infeasible for accurate estimation. As more measurements are taken, the posterior distribution is narrower and more accurate for $D=50$ than for $D=10$. However, $I_{\rm MINE}$ with BO due to the scalability issue and computational challenge still shows a relatively high variance at $D=100$, while $\texttt{SAGABED}$ and gradient-based method estimate the parameters more precisely and accurately. The resulting difference in parameter estimation exactly maps the difference in the MI lower bound estimator.  

\begin{table*}[!h] 
\footnotesize
\centering
\caption{Estimating mean and standard deviation of the posterior samples of the model parameters $\bm \theta$ using optimal designs $\bm d^*$ and real data observation $\bm y^*$ (use $\bm \theta_{\textup{true}} = [1,4]$ to generate $\bm y^*$)}
\label{tab:noise}
\begin{tabular}{@{}ccccccccc@{}}
\toprule
Method      & \multicolumn{2}{c}{D=1}    & \multicolumn{2}{c}{D=10} & \multicolumn{2}{c}{D=50}  & \multicolumn{2}{c}{D=100} \\ \midrule
& $\hat{\theta}_1$      & $\hat{\theta}_2$  & $\hat{\theta}_1$      & $\hat{\theta}_2$      & $\hat{\theta}_1$      & $\hat{\theta}_2$     & $\hat{\theta}_1$      & $\hat{\theta}_2$     \\
MINE+SGD &-1.39$\pm$2.54 & 6.03$\pm$\bf 0.93  & 0.51$\pm$\bf0.44 & 2.99$\pm$0.67 & 1.20$\pm$0.18 & 3.79$\pm$0.23  & 0.97$\pm$0.05   & 4.04$\pm$0.04  \\
MINE+BO  & -1.42$\pm${\bf0.81} & 2.98$\pm$1.19 & 1.22$\pm$0.58 & 4.93$\pm$0.91 & 0.71$\pm$0.25   & 3.66$\pm$0.40 & 1.35$\pm$0.21   & 4.79$\pm$0.26  \\
SMILE+GES & 2.76$\pm${1.36} & 5.74$\pm$3.08 & 0.83$\pm$0.56   & 4.69$\pm${\bf0.58} & 1.11$\pm${\bf0.13}   & 4.25$\pm${\bf0.19}  & 1.02$\pm${\bf0.04}   & 3.98$\pm${\bf0.03}  \\ \bottomrule
\end{tabular}
\vspace{-0.4cm}
\end{table*}

\subsection{ Pharmacokinetic Model}
Next we illustrate our approach for the design problem of determining the optimal measurements of blood sampling times for a pharmacokinetic (PK) study, which basically involves the administration of a drug to an individual and then investigate the drug distribution, absorption and elimination at certain times to analyze the underlying kinetics. The computational model introduced by \citet{ryan2014towards} is used to simulate the drug concentration at a specific time. 
There are three governed model parameters: the volume of distribution $V$, the absorption rate $k_a$ and the elimination rate $k_e$. The PK model is defined by a latent variable for the drug concentration at design time $t$, 
\begin{equation}
    z(t) = \frac{D_v}{V}\frac{k_a}{k_a-k_e} \left(e^{-k_e t} - e^{-k_a t}\right) (1+\epsilon_{1t}) + \epsilon_{2t}, \label{eq:pk}
\end{equation}
where $k_a>k_e$ is a constraint, $D_v=400$ is a single fixed dose at the beginning of the experiment, $t \in [0,24]$ is the design domain, $\epsilon_{1t} \sim \mathcal{N}(0, 0.01)$ and $\epsilon_{2t} \sim \mathcal{N}(0, 0.1)$ are multiplicative noisy sources that are often observed in the drug data. Note that Eq.~\eqref{eq:pk} is a simplified model and we assumes that the design parameters $\bm \theta = (V, k_a, k_e)$ are the same for the group of patients. As suggested by \citet{ryan2014towards}, the prior distribution for $\bm \theta$ is given by
\begin{equation}
    \log (\bm \theta) \sim \mathcal{N} \left( \left[ \begin{matrix} \log 20 \\ \log 1 \\ \log 0.1 \end{matrix}\right], \left[ \begin{matrix} 0.05 & 0 & 0 \\ 0 & 0.05 & 0 \\ 0 & 0 & 0.05 \end{matrix}\right] \right) \label{eq:pk_prior}
\end{equation}

In this example, we assume to take $D$ measurements that mean we have a group of $D$ patients but take only one blood sample from each patient at time $t$. Thus the design parameters are $\bm \xi = [t_1, ...,t_D]^T$ and the corresponding observations are $\bm y = [y_1, ...,y_D]^T$. It is noted that the PK model in Eq.~\eqref{eq:pk} allows us to analytically derive the pathwise gradients in terms of the sampling path \citep{kleinegesse2020bayesian}
\begin{equation}
    \frac{\partial z}{\partial t} = \frac{D}{V}\frac{k_a}{k_a-k_e} \left( k_a e^{-k_a t}-k_e e^{-k_e t} \right) (1+\epsilon_{1t}), \label{eq:pk_grad}
\end{equation}
The available gradients in Eq~\eqref{eq:pk_grad} enable to find the optimal design of blood sampling times using gradient-based methods, which can be considered as a baseline method. 

Instead of the simpler case by setting $D=1$ in the linear example, we initially set $D=10$ here and randomly generate 10 design time in $t \in [0, 24]$ hours as the initial state. Then 10,000 model parameters samples $\bm \theta^{(i)}$ are drawn from the prior distribution in Eq.~\eqref{eq:pk_prior}. Using these samples, we can collect the corresponding data samples $\bm y^{(i)}$ using the current design $t$. For neural network model $\mathcal{T}_{\bm \psi}(\bm \theta, \bm y)$, we use one hidden layer of 100 neurons with \texttt{ReLU} activation function as well as \texttt{Adam} optimizer with learning rates $\ell_{\psi}=10^{-4}$ and $\ell_{\xi} = 10^{-2}$. Hyperparameters tuning details can be found in the supplementary material. 

\begin{figure}[h!]
    \centering
    \includegraphics[width=0.22\textwidth]{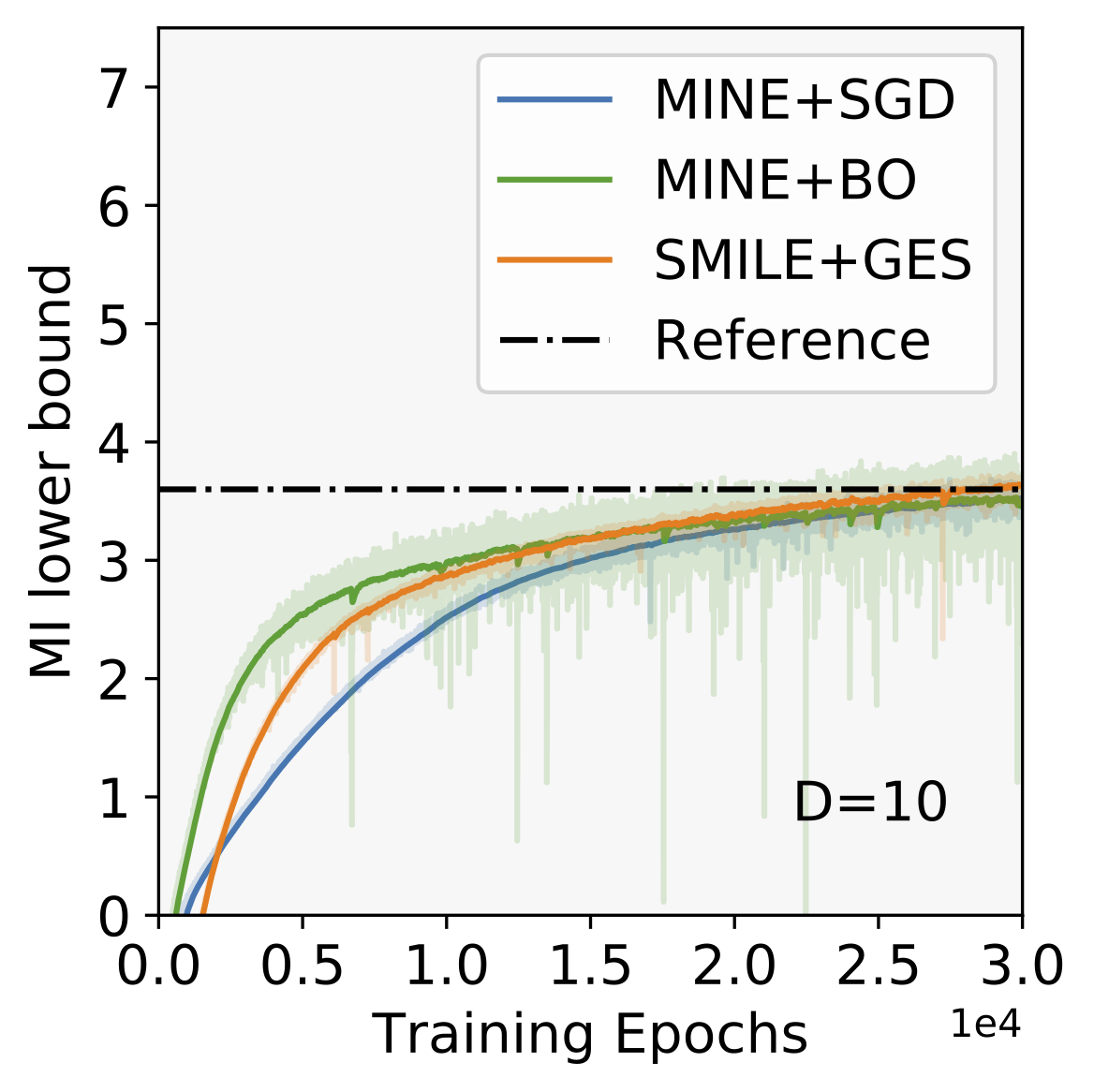}
    \includegraphics[width=0.22\textwidth]{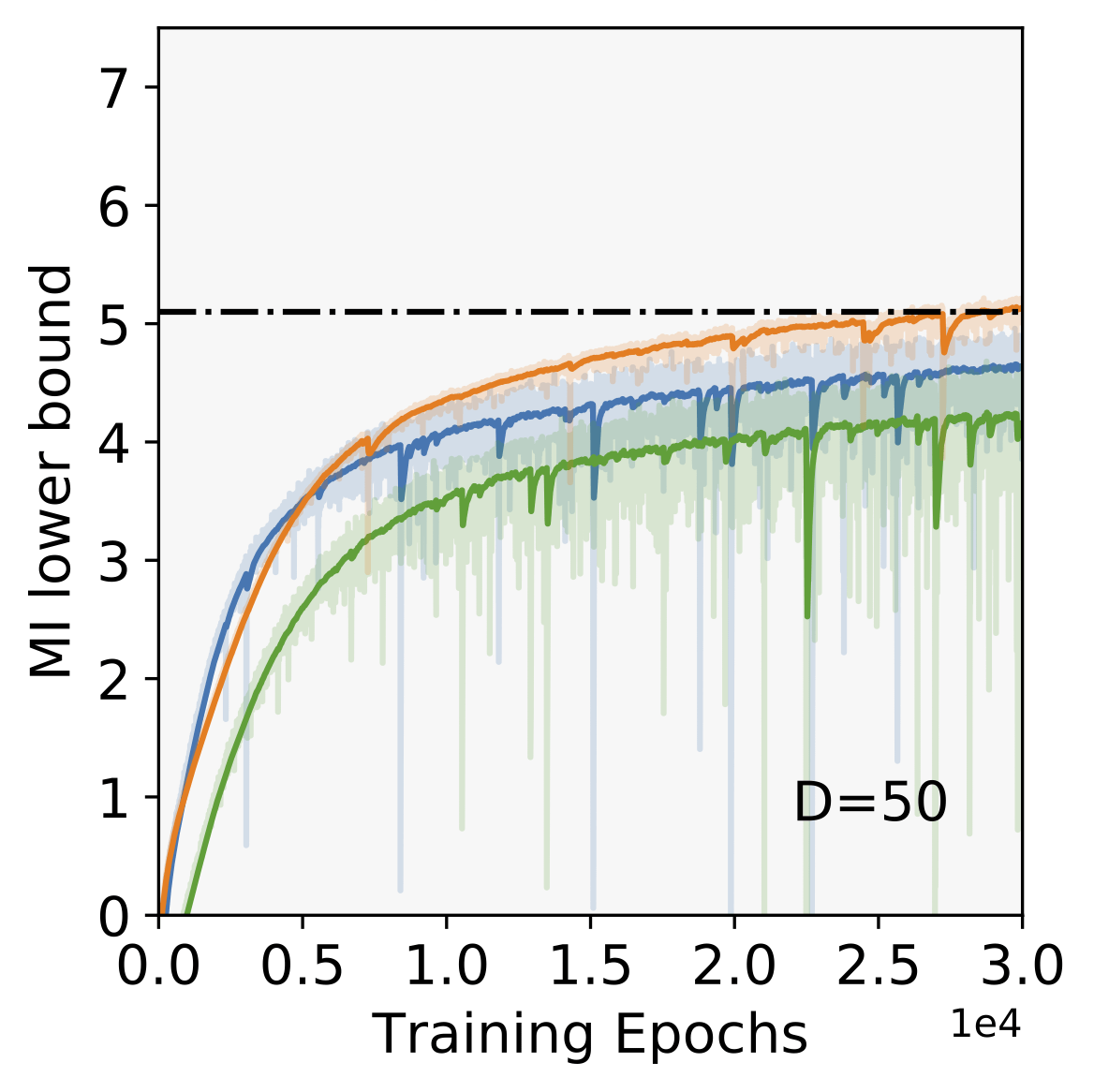}
    \includegraphics[width=0.22\textwidth]{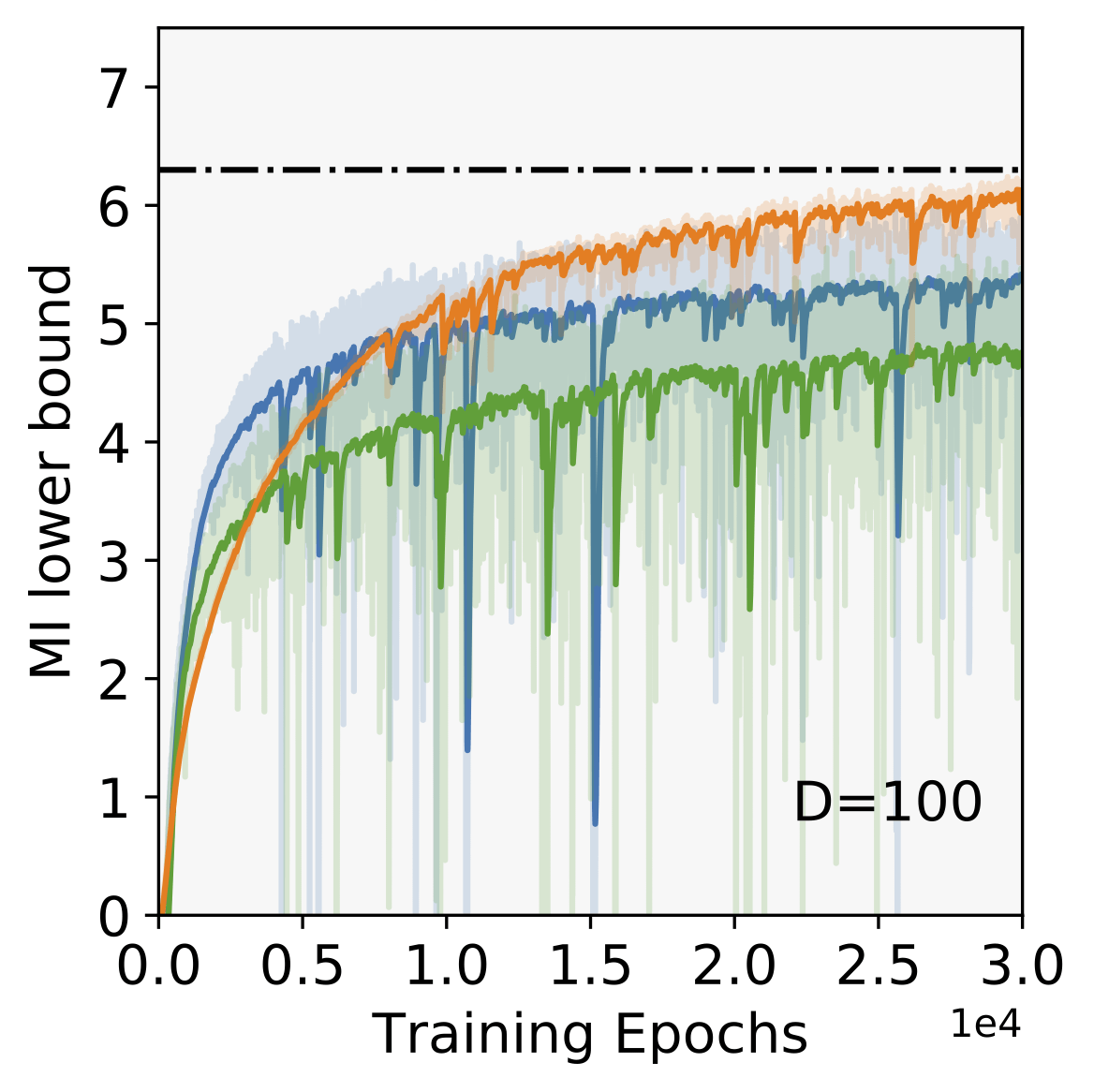}
    \includegraphics[width=0.22\textwidth]{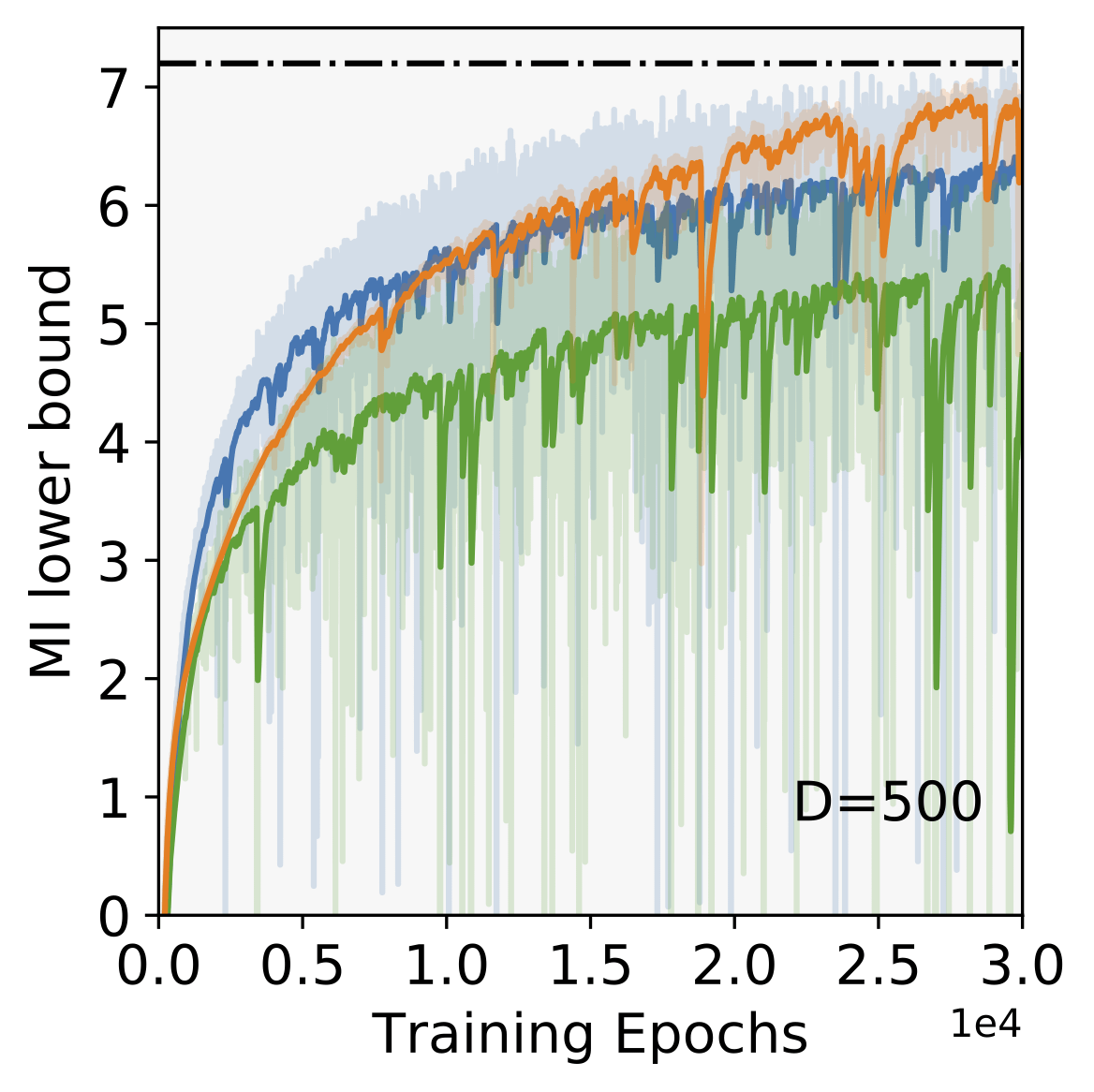}
    \caption{MI lower bound as a function of neural network training epochs for different number of measurements: $D$=10, 50, 100 and 500 in the study of PK model.}
    \label{fig:pk_mi}
\end{figure}

Figure \ref{fig:pk_mi} shows the MI lower bound as a function of the neural network training epochs in different sampling times $D=10,50,100$ and $500$ for the PK study. We compare the {\texttt{SAGABED}} (refers SMILE+GES in Figure \ref{fig:pk_mi}) with the baselines that use MINE with SGD and BO optimizer respectively. When $D=10$, the MI lower bound using \texttt{SAGABED} converges to around 3.6 that is close to a reference MI value at the optimal design $t^*$. The baselines show a comparative performance compared to {\texttt{SAGABED}}. As more measurements $D=50$ are performed, all three methods converge to a higher MI lower bound; this is as expected since more information are obtained from more data of model parameters. {\texttt{SAGABED}} shows a negligibly small bias compared to the reference MI value and has a low variance, which outperforms the other two baselines. 

We now focus on the high dimensional cases, e.g., $D=100$ and 500, that are more challenging settings. Here we use a 5 layered neural network with 50 neurons for each layer instead of the neural network architectures in the low dimensional cases. While {\texttt{SAGABED}} shows a larger bias of the final MI lower bound compared to the reference value due to increasing dimensions, it is still reasonably acceptable and higher than the baselines. In the high dimensional setting, the gradient-based method increases fast initially but it is easy to be trapped into local minima; the BO method has no this issue but does not scale to such high dimensionality. Also, the MI lower bound estimated by $I_{\rm MINE}$ shows a high variance whereas $I_{\rm SMILE}$ has a lower variance across all different tasks. That is an important feature that may potentially reduce the variance of final optimized design and also avoid additional efforts on optimizing hyperparameters, e.g., learning rates.

\begin{figure}[h!]
    \centering
    \includegraphics[width=0.5\textwidth]{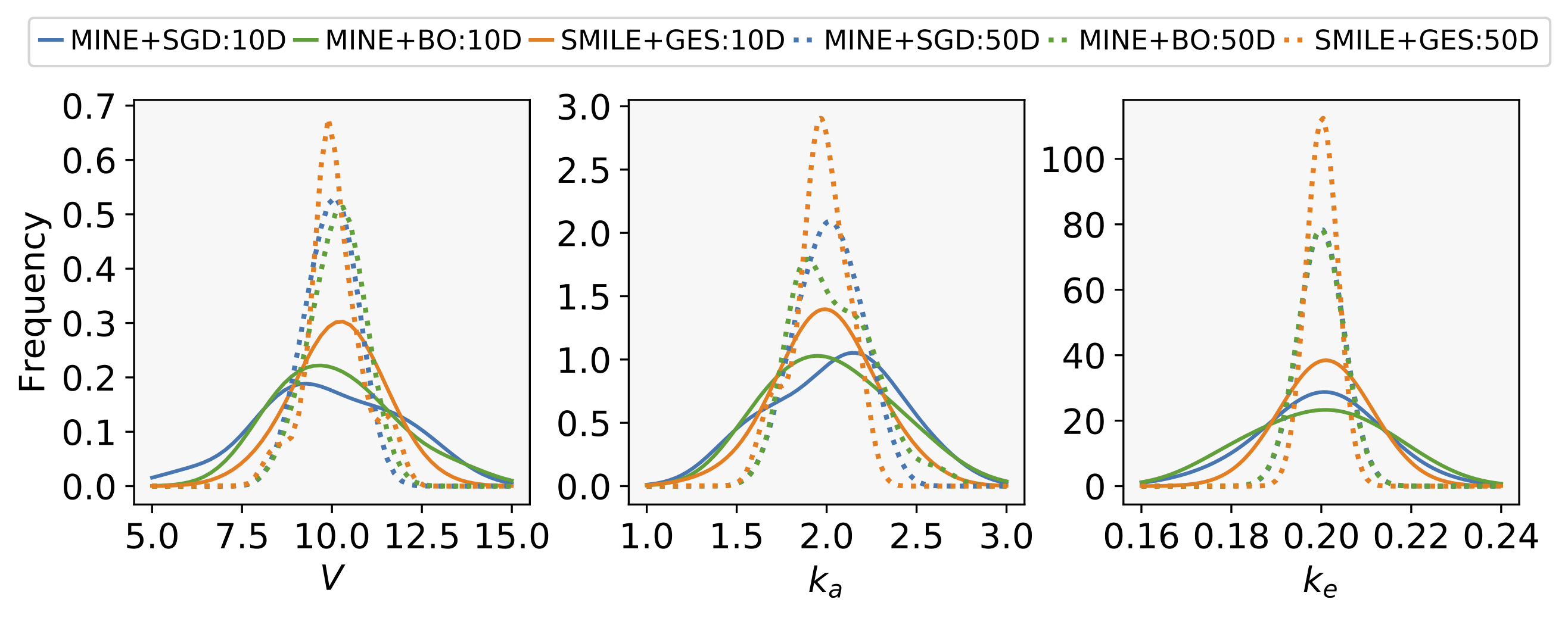}
    \caption{Marginal posterior distributions of the model parameters for $D=10$ (solid curves) and $D=50$ (dotted curves). Kernel density estimate is used to approximate the posterior samples. \texttt{SAGABED} shows a much narrower posterior distribution of the model parameters.}
    \label{fig:pk_posterior}
\end{figure}
Once the optimal design $t^*$ is obtained, we can conduct the experiment to generate the real-world data $\bm y^*$ by assuming a true model parameter $\theta_{\rm true} = [20, 2.0, 0.2]$. Then we compute the posterior density $p(\bm \theta| t,^*, y^*)$ using the learned neural network $\mathcal{T}_{\bm \psi^*} (\bm \theta, \bm y)$, and obtain the posterior samples using MCMC sampling. Figure \ref{fig:pk_posterior} shows the marginal posterior distribution of each model parameters. Using $I_{\rm SMILE}$, we achieve a more accurate and much narrower estimate of all the model parameters with a lower variance than the baseline methods. 

\subsection{Tuning for Quantum Control}
Typically, the reliable capability to manipulate qbit states is critical to quantum technology. For instance, radio-frequency pulses can be used to change of state of spin-up and spin-down, and patients benefit from MRI scans that use these approaches. In this example, we aim to conduct \texttt{SAGABED} to simulate a tuning process so that we can actively control the desired duration and frequency of pulses to flip electron spins in a reliable scheme. Specifically, the model parameters $\bm \theta = (\theta_1, \theta_2)$ are Rabi frequency and true center of resonance relative to the reference frequency, and the design parameters $\bm \xi = (t, \Delta f)$ are the duration time of microwave pulse and the detuning relative to a reference frequency. We here use the physical simulator developed by \citet{nistobed} to simulate the tuning process for quantum control. The physical simulator used here can be interpreted as an implicit model where the gradient can not be computed exactly. We assume that our measurements are subject to Gaussian noise, $\mathcal{N}(0,1)$. The simulation domain is discretized by 101$\times$101 grids and a finer grid would lead to a more accurate prediction but at a higher computational cost. The design domain is set to $t \in [0,1]$ and $\Delta f\ \in [-10, 10]$ and we therefore assume uniform prior distributions $p(t) = U(0,1)$ and $p(\Delta_f) = U(-10,10)$. 

\begin{figure}[h!]
    \centering
    \includegraphics[width=0.48\textwidth]{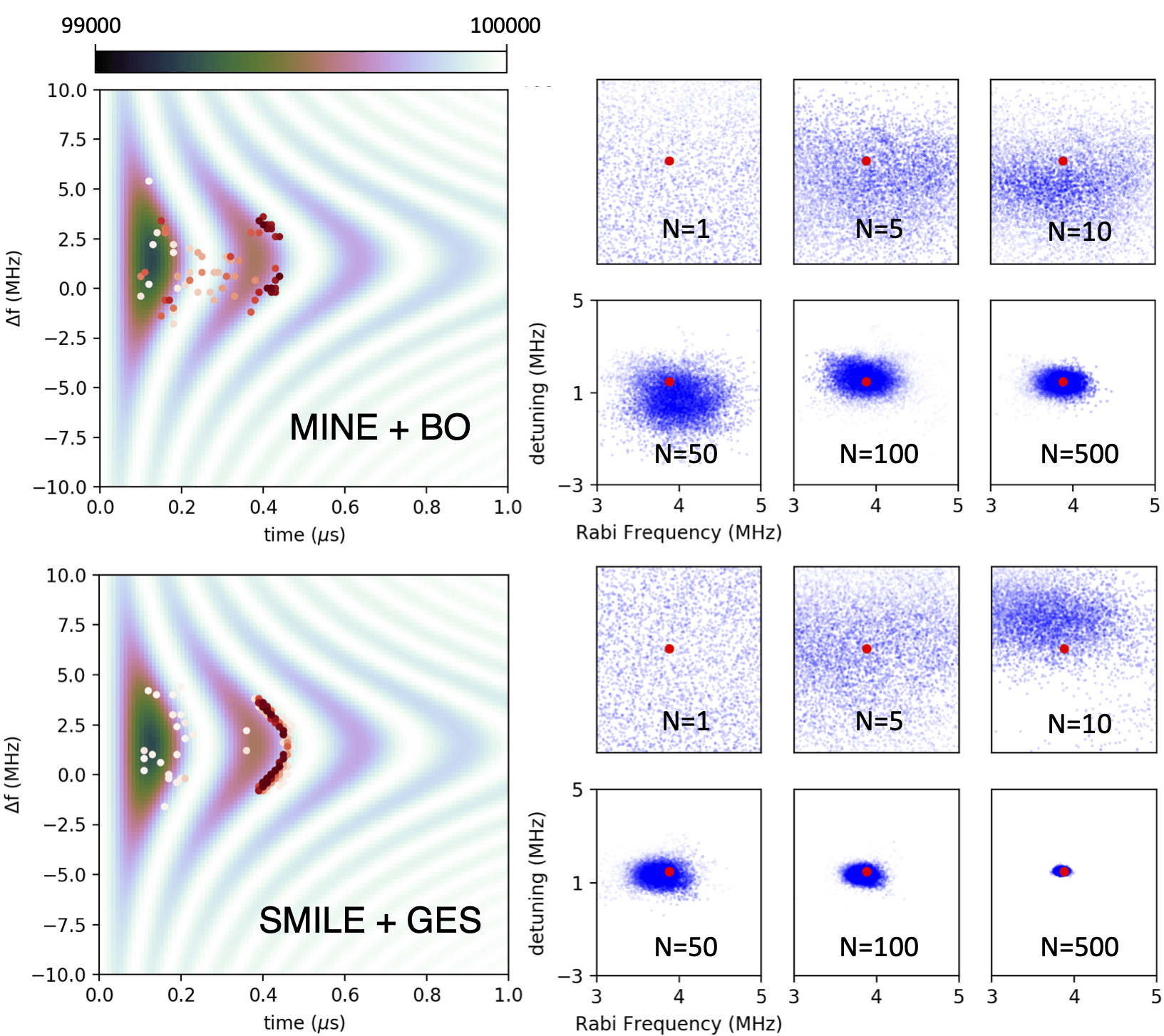}
        \caption{Performance comparison between \texttt{SAGABED} and BO method for tuning quantum pulse. The contour image (left column) shows the model photon counts for optically detected spin manipulation for pulse duration and amounts of detuning from the spin's natural resonance frequency. The right column displays the evolution of the posterior distribution with the number of designed measurements. The red points are the true model parameters. 
    }
    
    \label{fig:quantum}
\end{figure}

We use a one-layer neural network with 200 neurons and a \texttt{ReLU} activate function as well as \texttt{Adam} optimizer with learning rates $\ell_{\psi}=10^{-3}$ and $\ell_{\xi} = 10^{-2}$. We still start from a simple case, i.e., only one measurement $N=1$, and then gradually increase the number of measurements to $N=5, 10, 50, 100$, and 500 finally. We randomly draw 10,000 uniform prior samples $\bm \theta^{(i)}$ and then simulate 10,000 corresponding samples of the outcome $\bm y$ that is the Rabi counts. The optimal design results are shown in the left column in Figure \ref{fig:quantum}. Once we have done the neural network training, we assume the true model parameters $\bm \theta_{\rm true} =[3.85, 1.67]$ as the true Rabi and detuning frequency. As similar to the previous examples, we can conduct a real-world measurement of the tuning control process given the optimal design $\bm \xi^*$ and learned neural network $\mathcal{T}_{\bm \psi^*}(\bm \theta, \bm y)$. As a result, the posterior samples are obtained and illustrated by the right column in Figure \ref{fig:quantum}. 

Our main purpose in this example is to demonstrate the superior performance of \texttt{SAGABED} on the robustness of the optimal design and their effect on the variance of the posterior samples. As a comparison, BED with the BO method shows a similar performance on the low-dimensional cases but a high variance when the dimensionality increases to 100. Even more design are collected, the posterior distribution in the BO method is difficult to be narrowed, as shown in Figure \ref{fig:quantum}. On the contrary, \texttt{SAGABED} demonstrates a faster narrowing rate to the true model parameters and it displays a much lower variance of the posterior samples in the high-dimensional design problems. This also illustrates the significant advantages of \texttt{SAGABED} in addressing the scalability challenges compared to the BO method. 

\section{Conclusion}

In this paper, we develop a general unified framework that utilizes the stochastic approximate gradient for BED with implicit models. Without the requirement or assumption of pathwise gradients, our approach allows the optimization to be carried out by stochastic gradient ascent algorithms and therefore scaled to substantial high dimensional design problems. Several experiments demonstrate that our approach outperforms the baseline methods, and significantly improves the scalability of BED in high dimensional settings.

The future work will focus on the extension of our proposed framework to sequential Bayesian experimental design (SBED) that involves an iterative update of the prior and posterior distribution. We plan to improve the computational efficiency of the \texttt{SAGABED}, which may play a more critical role in the SBED scenario. We are also interested in improving the stochastic approximate gradient methods and enhancing the global exploration capability to escape the sub-optimal design caused by the issue of local minima in current SGA algorithms, e.g., standard and guided ES \citep{zhang2020scalable,zhang2021directional}.  


\subsubsection*{Acknowledgements}
This work was supported by the U.S. Department of Energy, Office of Science, Office of Advanced Scientific Computing Research (ASCR), Applied Mathematics program; and by the Artificial Intelligence Initiative at the Oak Ridge National Laboratory (ORNL). This work used resources of the Oak Ridge Leadership Computing Facility, which is supported by the Office of Science of the U.S. Department of Energy under Contract No. DE-AC05-00OR22725.

\bibliographystyle{plainnat}
\bibliography{reference.bib}

\end{document}